%% file: root.tex
\documentclass[letterpaper, 10 pt, conference]{ieeeconf}  

\IEEEoverridecommandlockouts                              

\usepackage{amsmath} 
\usepackage{amssymb}  
\usepackage{graphicx}
\usepackage{balance}
\usepackage{wrapfig}
\usepackage{lipsum}

\usepackage{subcaption}
\usepackage{xcolor}
\usepackage{hyperref}
\usepackage{cleveref}
\usepackage{amsmath} 
\usepackage{amssymb}  

\usepackage{esvect}
\usepackage{algorithm,algpseudocode}
\usepackage{amssymb} 
\usepackage{multirow}
\usepackage{float}
\usepackage{adjustbox}
\usepackage{booktabs}

\definecolor{nscolor}{rgb}{0.188, 0.478, 0.624}

\algrenewcommand{\algorithmiccomment}[1]{// #1}

\newcommand{\methodfullname}{
Feedback Matters: Augmenting Autonomous Dissection with Visual and Topological Feedback
}

\definecolor{darkgreen}{RGB}{0,150,0}
\newcommand{\uparrowgreen}{\textcolor{darkgreen}{\textuparrow}}
\newcommand{\downarrowgreen}{\textcolor{darkgreen}{\textdownarrow}}

\title{\LARGE
\methodfullname
}

\author{Chung-Pang Wang$^*{}^1$, Changwei Chen$^*{}^1$, Xiao Liang$^1$, Soofiyan Atar$^1$, Florian Richter$^1$, Michael Yip$^1$ 
\thanks{$^*$These authors contributed equally to this paper.}
\thanks{$^{1}$ Department of Electrical and Computer Engineering, University of California San Diego, La Jolla, CA 92093, USA {\tt\small \{chw120, chc165, x5liang, satar, frichter, yip\}@ucsd.edu}}%
}

\begin{document}

\maketitle
\thispagestyle{empty}
\pagestyle{empty}

\begin{abstract}

   Autonomous surgical systems must adapt to highly dynamic environments where tissue properties and visual cues evolve rapidly. Central to such adaptability is feedback: the ability to sense, interpret, and respond to changes during execution. While feedback mechanisms have been explored in surgical robotics, ranging from tool and tissue tracking to error detection, existing methods remain limited in handling the topological and perceptual challenges of tissue dissection. In this work, we propose a feedback-enabled framework for autonomous tissue dissection that explicitly reasons about topological changes from endoscopic images after each dissection action. This structured feedback guides subsequent actions, enabling the system to localize dissection progress and adapt policies online. To improve the reliability of such feedback, we introduce visibility metrics that quantify tissue exposure and formulate optimal controller designs that actively manipulate tissue to maximize visibility. Finally, we integrate these feedback mechanisms with both planning-based and learning-based dissection methods, and demonstrate experimentally that they significantly enhance autonomy, reduce errors, and improve robustness in complex surgical scenarios.

\end{abstract}

\input{tex_files/introl_and_related_work}

\input{tex_files/method}

\input{tex_files/experiment}
\bibliographystyle{IEEEtran}
\bibliography{root}

\end{document}

%% file: tex_files/introl_and_related_work.tex
\section{Introduction}


Learning and adaptation are impossible without observational feedback. 
In human cognition, feedback is the core mechanism for mastering skills: from infants adjusting their first steps after a stumble, to expert surgeons refining their skill based on subtle visual cues. 
Robotic autonomy similarly depends on feedback, where sensing the environment and responding to unexpected changes are essential for robust and adaptive behavior. 
Surgical robotics is no different especially for sub-tasks that demands high precision and reactive capability to complex nonlinear changes, such as tissue dissection and suturing. 
Yet current surgical automation works lacks the reliable feedback-enabled autonomy required to detect and react to failures in real tissue environments.

Feedback mechanisms have been explored in surgical robotics to enable more adaptive and reliable autonomy. 
A prominent line of work has focused on tool tracking \cite{lucas_Differentiable_tool_tracking, 10610378} with the da Vinci Research Kit (dVRK), which has enabled applications such as visual servoing of surgical instruments \cite{medic_ieee}. 
Similarly, needle tracking \cite{chiu2022real} has been investigated to facilitate suturing subtasks, including the delicate hand-off of the needle between instruments.
Tissue modeling and tracking \cite{10610263,kim2025srt} from real-world sensor data provide dynamic estimates of tissue deformation and support safe manipulation \cite{shinde2024jiggle} and retraction \cite{medic_arxiv}. 
Other feedback mechanisms address error detection, such as methods for blood flow detection that leverage physics models and visual cues to identify bleeding events and guide blood suction \cite{richter2021autonomous}. 
More recently, learning-based methods \cite{11027660, kim2024surgical} have emerged for their improved precision and generalization ability in the presence of occlusion, noise, and modeling error. Learned policy internalizing sensor feedback through data-driven optimization.
However, they lack explainability and theoretical guarantees, and their performance often degrades and produce unpredictable failures in the absence of large scale datasets, limiting it's reliability in surgical settings. 
\begin{figure}[t]
    \centering
    \includegraphics[width=\linewidth]{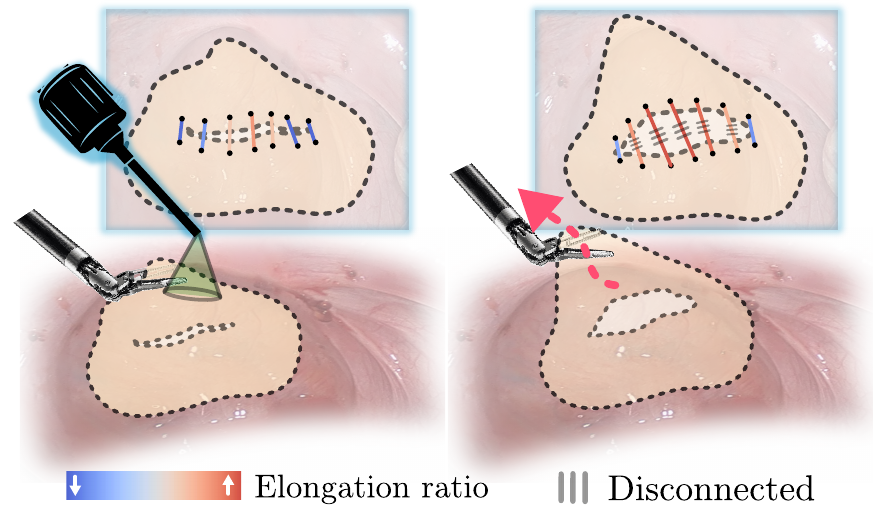}
    \caption{Illustration of the importance of exposure maximization for tissue dissection outcome verification. When tissue is dissected, an estimation method is uncertain about tissue connectivity condition due to lack of stretching. A proper controller should not only stretch the dissection site, but also consider its visibility under the camera observation model.}
    \label{fig:reveal_fig}\vspace{-1.5em}
\end{figure}

Despite progress in feedback mechanisms for surgical automation, relatively little attention has been given to the high-precision surgical task of tissue dissection. 
Previously, several works have attempted to automate dissection through supervised controllers for precise 3D incisions on porcine tissues \cite{8884158, 10.1007/978-3-030-32239-7_36}, frameworks for boundary detection and tool tracking \cite{oh2023framework}, and customized systems for tumor resection \cite{ge2021supervised,10354422}. 
Reinforcement learning has also been applied to automate electrocautery dissection from point cloud input \cite{karimi2024reward}, while other efforts target sensing and reconstruction of dissection sites \cite{marahrens2024ultrasound,franccois2021image}. 
In parallel, simulation-based research has advanced modeling of tissue dissection \cite{ge2024enhancing,ou2025cressim}, providing important platforms for testing and training autonomous surgical systems.
Yet, despite their contributions, none of these methods have explored incorporating robust feedback mechanisms. 
They automate execution given predefined goals but cannot verify task completion or adapt when dissection progresses differently than expected, which is a critical shortcoming in real surgical environments.
In addition, many critical feedback signals are not directly observable and must be actively maximize exposure through manipulation—for example, stretching tissue to verify that a cut has been fully completed.
This tight coupling between perception and action underscores the need for feedback mechanisms that go beyond passively interpreting sensor data. 
Instead, they must also drive active interventions that expose hidden state information essential for safe, accurate, and autonomous dissection.


To address these challenges, we propose a feedback-enabled framework for autonomous tissue dissection. 
Our method explicitly reasons about topological changes of tissue from endoscopic image data after each dissection action, providing structured feedback to subsequent steps of an autonomous dissection agent. 
Furthermore, we introduce two visibility metrics of dissected areas, which quantify the reliability of the feedback based on tissue surface exposure to the camera and can be integrated into a dissection agent to improve the quality of the feedback. 
Finally, through extensive experiments, we demonstrate the versatility of our visual and topological feedback for tissue by integrating it with both planning-based and learning-based dissection methods, analyzing their limitations, and showing experimentally how feedback improves autonomy in each case.

%% file: tex_files/method.tex
\section{PRELIMINARIES}
\begin{figure*}[t]
    \centering
    \includegraphics[width=0.95\linewidth]{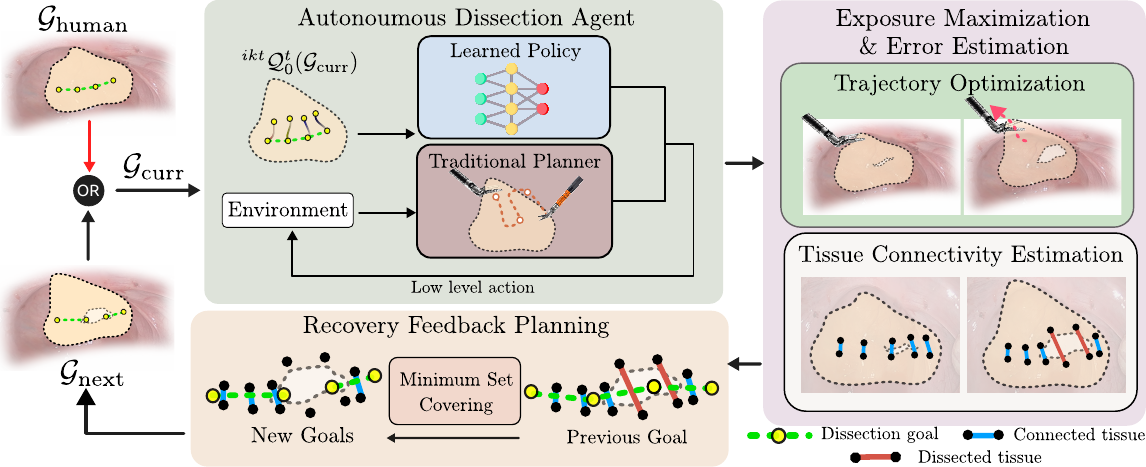}
    \caption{\textit{\textbf{Our Autonomous Dissection Pipeline.}} Our framework operates in a feedback loop. The process begins with a human-provided dissection goal $\mathcal{G}_{\text{human}}$, which the autonomous agent executes. Next, an exposure maximization controller manipulates the tissue to improve visibility for the subsequent error estimation. Finally, a recovery planner estimates the remaining tissue connectivity. If the dissection is incomplete, it generates a corrective goal, and the loop repeats until the task is successfully completed or a maximum number of attempts is reached.}
    \label{fig:pipeline}\vspace{-1.5em}
\end{figure*}

\textbf{Deformable Tissue Simulation:}
Extended Position-based Dynamics (XPBD) simulator has been used as dynamic models for tissue manipulation in previous works \cite{medic_ieee}. 
This work adopts it as the underlying dynamic model for exposure maximization control.
XPBD represents a system with particles $\mathbf{x}_t$, robot control $u_t$ and geometric constraints, iteratively solving particle states for constraint satisfaction. 
The dynamic equation can be expressed as
\begin{equation}
\begin{split}
        \mathbf{x}_{t+1} &= f_\text{xpbd}(\mathbf{x}_t, u_{t})
\end{split}
\end{equation}
A piece of tissue is represented a triangular mesh $\mathbf{M}: \{\mathbf{v}, \mathbf{F}\}$, where $\mathbf{v}$ is a set of mesh vertices and $\mathbf{F}: \{F_1,...,F_b\}$ denotes triangle faces.
The tissue is simulated with vertices as particles, and with distance constraints defined on its edges, bending constraints defined on neighboring faces. 
The robot control $\mathbf{u}_{t} \in \mathbb{R}^3$ can achieve grasping and tissue retraction effect by simulating it as a boundary condition on position offsets of certain tissue particles. 

\textbf{Tissue tracking:}
Tissue tracking is an important technique for localizing and tracking anatomy over time. Tissue tracking methods are denoted as 
\begin{equation}
    p_j = \mathcal{Q}_i^j(p_i)
    \label{eqn:tracking}
\end{equation}
where $i, j$ indicates time stamps or frames. The tracking method takes in coordinate query $p_i$ at frame $i$ and produces the corresponding coordinate $p_j$ at frame $j$, describing the movement of $p$.
In this work, we consider two of the most widely used approaches for tissue tracking: image keypoint tracking (IKT) ${}^{ikt}\mathcal{Q}$ and model-based tracking (MBT) ${}^{mbt}\mathcal{Q}$. While both methods track point in the image frame (i.e. $\mathcal{Q}: \mathbb{R}^2 \mapsto \mathbb{R}^2$), their tracking mechanisms are different. IKT tracks points purely with flow with image feature, such as \cite{cotracker}. MBT tracks point based on a physics model's prediction, in this case, XPBD model. The full tracking procedures are defined as follow:
\begin{equation}
    \begin{split}
      {}^{mbt}\mathcal{Q}_i^j(p_i, \mathbf{x}_{i:j}):\   & \hat p_i = \text{raycast}(p_i, \mathbf{x}_i, \mathbf{M}, C)\\
        & k, \alpha, \beta, \gamma = \text{barycentric}(\hat p, \mathbf{x}_i, \mathbf{M})\\
        &\hat p_j = \text{track}(\mathbf{x}_j, k, \alpha, \beta, \gamma)\\
        & p_j = \text{project}(\hat p_j, C).
    \end{split}
    \label{eq:model_based_tracking}
\end{equation}
In above equations, $C$ represents a known model-view matrix. 
These procedure includes (1) performing ray-casting of an image point to back-project it onto the mesh surface, (2) using its barycentric coordinates to track the point after mesh deform, (3) project the 3D point back to the image frame. 
For simplicity, dependency of ${}^{ikt}\mathcal{Q}$ on image inputs, and ${}^{mbt}\mathcal{Q}$ on the mesh and camera model are omitted.



\section{METHODS}

The proposed method considers a task where a dissection goal is provided as an input by expert surgeons, and it aims at removing tissue connectivity around the given target.
A dissection goal is parameterized as a polyline $\mathcal{G}^{(t)} \;=\; \bigl\{\, p^{k}_t \in \mathbb{Z}^2 \;\bigm|\;k=1,\dots,M \bigr\}$ that lies on the pixel space on an endoscopic image.
This representation is chosen as endoscopic displays are the most common interaction modality. 
We assume the dissection goal to be a low-curvature path and we denote $\vec{\mathbf{n}}_{t}$, $\vec{\mathbf{m}}_{t}$ to be the orthogonal and parallel unit vectors of the dissection goal $\mathcal{G}^{(t)}$.
An intuitive understanding of tissue connectivity removal is that tissue connections do not intersect with the dissection polyline, $\mathcal{G}$. 
Its exact definition and metrics is detailed in \autoref{sec:detection}.

This work particular concerns the completeness of autonomous tissue dissection, which is the challenge of ensuring that all targeted tissue is fully severed, leaving no remaining connections.
To address this challenge, an autonomous dissection pipeline shown in \autoref{fig:pipeline} is introduced, that has components of (1). autonomous dissection agents including both planning and learning-based methods (detailed in \autoref{sec:policies}), (2). trajectory optimization for exposure maximization (\autoref{sec:exposure_control}), and (3). recovery planning that proposes future actions to achieve completeness (\autoref{sec:feedback}). 
Three components are designed to execute iteratively, providing actionable feedback to surgical dissection task.

\subsection{Autonomous Dissection}\label{sec:policies}
We achieve autonomous dissection through two distinct paradigms: (1) a learned policy that executes bimanual retraction and cutting with visual inputs (2) a planning-based dissection agent that follows a 3D path generated by projecting tracked image keypoints into the robot's workspace.

\noindent \textbf{3D Visuomotor Dissection Policy.} Depth-based representations including point clouds from stereo matching and monocular depth are used as observations of the tissue.
Because the state-of-the-art disparity estimation algorithm \cite{stereo_anywhere} prioritize accuracy over inference speed,
we adopt a slow-fast strategy: high-frequency left monocular depth image, $\mathbf{M}^{L}_{t}\in\mathbb{R}^{3 \times H \times W}$, estimated by Video Depth Anything \cite{video_depth_anything}, and lower-frequency point clouds, $\mathbf{P}_t \in \mathbb{R}^{N\times3}$, reconstructed from a modified Stereo Anywhere \cite{stereo_anywhere}, in which Depth Anything v2 \cite{depth_anything_v2} is replaced  with Video Depth Anything for monocular depth estimation in Stereo Anywhere for more consistent deep stereo matching.

We trained our policy via behavioral cloning on the human demonstrations we collected, using action chunking with transformer (ACT) \cite{ACT} to generate a bi-manual action sequence given the current observations.
The policy is learned to retract the tissue and perform dissection autonomously given a dissection goal from human demonstrations. 

\noindent \textbf{Planning-based Dissection.} 
Given a tracked dissection goal $\mathcal{G}^{(t)}$ and a grasping point $p^{\text{grasp}} \in \mathbb{Z}^2$ on the image plane initially defined from human experts, the planner generates a sequence of actions. 
First, it creates a 3D path to grasp the tissue and then retracts along the goal's normal vector. 
The dissection trajectory is planned as a series of line segments, with the tool oriented along the path. 
Each cut is executed by moving to the target pose, inserting to a fixed depth, and closing the gripper to dissect.





\subsection{Tissue Connectivity Estimation}\label{sec:detection}
We combine keypoint tracking with a rule-based approach to estimate tissue connectivity conditions undergoing tissue deformations (e.g. maximizing exposure covered in the next section).
Defining the horizon of the estimation task as $t=t_d,\dots, t_e$, where $t_d$ is the time immediately after a dissection is executed, and $t_e$ denotes the end of the exposure and estimation procedure,
the initial keypoint set
$
\mathcal{K}^{(t_d)}=\{\mathbf{k}_i^{(t_d)}\}_{i=1}^{N_k}
$
is sampled as a uniform grid around the tracked dissection target $\mathcal{G}^t$. A temporary connectivity edge set $\mathcal{E} \subseteq \{\, e_{ij} \mid i,j \in \{1,2,\dots,N_k\},\; i \neq j \,\}$ is defined base on the grid, including an undirected edge for every two neighboring grid points. 

The tissue connectivity is estimated by iteratively computing the tracked keypoints temporal evolution,
\begin{equation}
\begin{aligned}
\mathcal{K}^{(t_e)} &= {}^{IKT}\mathcal{Q}_{t_d}^{\,t_e}\!\bigl(\mathcal{K}^{(t_d)}\bigr) \; ,
\end{aligned}
\end{equation}
and evaluating how much each edge's length has changed.
This elongation ratio is calculated for each edge as
\begin{equation}
    \begin{split}
        \rho(e_{ij}) &= \frac{\lVert \mathbf{k}_i^{(t_e)}-\mathbf{k}_j^{(t_e)}\rVert_2}{\lVert \mathbf{k}_i^{(t_d)}-\mathbf{k}_j^{(t_d)}\rVert_2}.\\
    \end{split}
\end{equation}
and an edge is viewed as disconnected (dissected) when it can be extended to $\tau$ times longer than its original length, i.e. $\rho(e_{ij}) \ge \tau$ where $\tau > 1$ is a ratio threshold. Finally, the connectivity set is updated to include only connected edges:
\begin{equation}
    \mathcal{E}_t = \{\, e_{ij}\in\mathcal{E} \mid \rho_{ij} < \tau \,\}.
\end{equation}




\subsection{Exposure Maximization Control}\label{sec:exposure_control}
The accuracy of our tissue connectivity estimation hinges on maximizing the elongation of tissue edges orthogonal to the dissection path. 
If this elongation is insufficient or poorly oriented, the geometric cues become ambiguous, making it difficult to reliably distinguish between connected and severed tissue as illustrated in \autoref{fig:reveal_fig}. 
Therefore, two factors are critical for a robust tissue connectivity estimation: the magnitude of the elongation and its visibility to the camera. 
To address this, we propose an Exposure Maximization Controller that actively deforms the tissue to create an optimal viewing configuration to ensure robust tissue connectivity estimation.
This controller solves an optimal control problem to find a robot action sequence that maximizes the visibility of the deformation. 

Let $\mathbf{M}_{t_d}$ be a surface mesh that is reconstructed from a segmentation and a point cloud \cite{liang2023real} after a dissection action.
We do not make assumptions on the connectivity of the tissue as it is the estimation objective.
Let $P_{t_d} \subset \mathbb{Z}^2 $ be the set of image pixels near the proximity of a given dissection target. 
We create a simulation $f_\text{xpbd}$ of $\mathbf{M}_{t_d}$, and formulate exposure maximization of $P_{t_d}$ as the following finite horizon optimal control problem:
\begin{equation}
\begin{aligned}
\min_{ \mathbf{u}_{t_d:t_e} } \quad & \sum_{\tilde{t}=t_d}^{t_e} \Big( \lambda_{\text{c}} \, \mathcal{L}_{\text{cam}}(P_{\tilde{t}}) + \lambda_{\text{d}} \, \mathcal{L}_{\text{deform}}(P_{\tilde{t}}) \Big) \\
\text{subject to} \quad & \mathbf{x}_{\tilde{t}+1} = f_{\text{xpbd}}(\mathbf{x}_{\tilde{t}}, u_{\tilde{t}}),\  P_{\tilde t} = {}^{mbt}\mathcal{Q}_{t_d}^{\tilde t}(P_{t_d}, \mathbf{x}_{t_d:{\tilde{t}}}), \\
& \mathbf{C}_{\text{vis}}(P_{\tilde{t}}) = 0 ,\ \tilde{t} = t_d, \ldots, t_e
\end{aligned}
\label{eq:ocp_formulation}
\end{equation}
where robot gripper position sequence $\mathbf{u}_{t_d:t_e} =\{u_{t_{d}},\ldots,u_{t_e}\}$ is initialized with the gripper position after dissection, 
$\mathcal{L}_{\text{cam}}, \mathcal{L}_{\text{deform}}$ and $\mathbf{C}_{\text{vis}}$ are the costs and constraints to guide the action to maximize visibility.
They are (1) camera-facing normal alignment loss to encourage surface normals to align with the viewing direction of the camera, (2) deformation expanding loss to encourage expansion of the dissected region, (3) a visibility constraint that enforces that $P_{\tilde t}$ are visible by the camera.

\textbf{Camera Facing Loss:}
This loss penalizes deviations of the surface normals from the camera viewing direction. Let $\hat P_{\tilde t}$ be the corresponding 3D points on the simulated mesh from raycasting $P_{\tilde t}$, and let $N_{\tilde t}$ be corresponding
surface normals.
The loss is defined as
\begin{equation}
    \mathcal{L}_{\text{cam}}(P_{\tilde t}) 
    = \sum_{\mathbf{n}_{i} \in N_{\tilde t}} \mathbf{n}_{i}^{\top} \mathbf{w}_c ,
    \label{eq:normal_loss}
\end{equation}
where $\mathbf{w}_c$ is the forward direction of the camera. 
The dot product $\mathbf{n}_{i}^{\top} \mathbf{w}_c$ is minimized when the vectors are antiparallel.
Minimizing $\mathcal{L}_{\text{cam}}$ encourages the dissection region to face toward the camera.

\textbf{Deformation Expanding Loss:}
We hypothesize that tissue expansion in the direction that is orthogonal to the dissection goal will likely lead to ideal elongation of uncertain tissue connections, thus assists the proposed estimation module. 
To achieve this, we start by computing the deformation gradient matrix over the dissection area $P_{t_d}$, for the ease of decomposing the deformation into orthogonal and parallel components.
The deformation gradient is computed as:
\begin{equation}
    \mathrm{F}^p_{\tilde t} = \mathbf{I} + \nabla_{p} \mathbf{d}^{p}_{\tilde t} = \begin{bmatrix} 1+\frac{\partial d^{(\tilde t)}_{x}}{\partial x} & \frac{\partial d^{(\tilde t)}_{x}}{\partial y} \\ \frac{\partial d^{(\tilde t)}_{y}}{\partial x} & 1+\frac{\partial d^{(\tilde t)}_{y}}{\partial y} \\
\end{bmatrix},\ p \in P_{t_d}
\end{equation} 
where $\mathbf{d}^{p}_{\tilde t}$ is the 2D displacement of pixel $p$ at time $\tilde{t}$ w.r.t $t_d$. Geometrically, the deformation gradient $\mathrm{F}^p_{\tilde t}$ transforms a vector from its initial to its deformed configuration through a sequence of pure stretch and rigid-body rotation.
Let $\vec{\mathbf{n}}_{t_d}$ and $\vec{\mathbf{m}}_{t_d}$ be unit vectors that are orthogonal and parallel to the dissection goal $\mathcal{G}^{(t_d)}$ at time $t_d$. 
Our goal here is to create a specific deformation behavior that expands the dissection region along $\vec{\mathbf{n}}_{t_d}$ and reduces the region's shearing along $\vec{\mathbf{m}}_{t_d}$. We achieve these objectives by minimizing the following function:
\begin{equation}\small
    \mathcal{L}_{\text{deform}}(P_{\tilde t})
    = \sum_{p \in P_{t_{d}}} -w_n(\vec{\mathbf{n}}_{t_d}^{\top}\mathrm{F}^p_{\tilde t}\vec{\mathbf{n}}_{t_d})^2 + w_s(\vec{\mathbf{n}}_{t_d}^{\top}\mathrm{F}^p_{\tilde t}\vec{\mathbf{m}}_{t_d})^2
    \label{eq:deform_loss}
\end{equation}
where $w_n$ and $w_s$ are the hyperparameter we tune. $\vec{\mathbf{n}}_{t_d}^{\top}\mathrm{F}^p_{\tilde t}\vec{\mathbf{n}}_{t_d}$ quantifies how much the deformation keeps $\vec{\mathbf{n}}_{t_d}$ aligned with itself and stretched along its own direction, and $\vec{\mathbf{n}}_{t_d}^{\top}\mathrm{F}^p_{\tilde t}\vec{\mathbf{m}}_{t_d}$ represents the loss of orthogonality under the deformation w.r.t the dissection goal.

We use a sampling-based optimal control solver for trajectory optimization. At each time step, we sample particles of future control trajectories with horizon $h$, simulating them in parallel and evaluating every particle with the proposed loss functions. To ensure visibility, we simply discard control sequences that leads to violation of the \textbf{visibility constraint}:
\begin{equation}\small
\begin{split}
    \mathbf{C}(P_{\tilde t + h}) =
    \begin{cases}
        1, & \text{if } \exists (u,v) \in P_{\tilde t + h} \;\; \text{s.t. } (u,v) \notin \Omega, \\[6pt]
        0, & \text{otherwise}
    \end{cases}\\
    \Omega = \{ (u,v) \in \mathbb{R}^2 \;\mid\; 0 \leq u < W,\; 0 \leq v < H \},
\end{split}
\end{equation}
where $W, H$ is the size of the endoscopic image.
A nominal control trajectory 
$\mathcal{\hat U}= \{\hat u_{\tilde t},\ldots,\hat u_{\tilde t+h} \}$
is obtained using the update rule from Model Predictive Path Integral (MPPI) \cite{williams2015model}, and we take a step forward in simulation with action $\hat u_{\tilde t + 1}$. This procedure is implemented using GPU accelerated language Warp \cite{warp2022} to ensure fast optimization.

\subsection{Recovery Feedback Planning}\label{sec:feedback}

In this section, a method that detects changes in tissue connectivity as error metric of tissue dissection outcome is first discussed. 
It is then followed by a recovery module that proposes next dissection targets to the dissection autonomy to remove previous errors.


\subsubsection{Next Targets Proposal}

\begin{algorithm}[t]
\caption{Greedy Next-Target Selection}
\label{alg:next-target}
\begin{algorithmic}[1]
\State \textbf{Input:} Uncut edges $\mathcal{E}^{\mathrm{uncut}}$, segment length $\ell$
\State \textbf{Output:} Next dissection target $\mathcal{G}_{\text{next}}$
\State Build candidate segments $S$
\State $U \gets \mathcal{E}^{\mathrm{uncut}}$ 
\State $\mathcal{G}_{\text{next}} \gets \varnothing$
\While{$U \neq \varnothing$}
  \State $s^\star \in 
        \arg\max_{s \in S}\;
        \Bigl\lvert \,\bigl\{\, e \in U \;\big|\; s \cap e \neq \varnothing \,\bigr\}\,\Bigr\rvert.$ \hfill 
  \State $\mathcal{G}_{\text{next}} \gets \mathcal{G}_{\text{next}} \cup \{ s^\star \}$
  \State $U \leftarrow U \setminus \bigl\{\, e \in U \;\big|\; s^* \cap e \neq \varnothing \,\bigr\}$
\EndWhile
\State \Return $\mathcal{G}_{\text{next}}$
\end{algorithmic}
\end{algorithm}

At the end of exposure maximization control ($t=t_d+t_e=T_{\text{end}}$), dissection along all segments in $\mathcal{G}^{(T_{\text{end}})}$ 
is successful if no remaining edge $e_{ij}\in\mathcal{E}_{T_{\text{end}}}$ intersects any target segment $g_m^{(T_{\text{end}})}$.
The intersecting remaining edges form the uncut set
\begin{equation}
    \mathcal{E}^{\mathrm{uncut}}
    := \{\, e_{ij}\in\mathcal{E}_{T_{\text{end}}} \mid \exists\, m:\ e_{ij}\cap g_m^{(T_{\text{end}})} \neq \varnothing \,\}.
\end{equation}

Assuming that the dissection policy/planner can reliably execute only a single cut segment of length at most $\ell$ per command,
we seek a minimum-size set of commanded segments that removes all remaining edges.
Let $\mathcal{G}_\text{next}$ be this set of command, the problem can be formulated as the following minimum set covering problem,
\begin{equation}
\begin{split}
&\min_{\mathcal{G}_{\text{next}}}\bigl|\mathcal{G}_{\text{next}}\bigr| \\
   \text{s.t.}\quad &
   \forall\, e\in\mathcal{E}^{\mathrm{uncut}},\ \exists\, s\in\mathcal{G}_{\text{next}}:\ s\cap e \neq \varnothing.
\end{split}
\end{equation}
This type of problem is known as NP-Hard, but can be solved using a greedy approximation algorithm as shown in \autoref{alg:next-target}.
The first step of the algorithm is construction a set of command candidates by sampling around $\mathcal{E}^{\mathrm{uncut}}$.
Each candidate $s(c, \theta)$ is a line segment parametrized with center $c$, and rotation angle $\theta$. The overall candidate set $S$ is defined as

\begin{equation}\small
\begin{split}
S = \bigl\{\, s(c,\theta) \;\big|\; 
   \exists\, e_{ij}\in\mathcal{E}^{\mathrm{uncut}} : d(e_{ij},c) < \hat d, \theta \in [0, \pi]
\bigr\}.
\end{split}
\end{equation}
where $d(\cdot)$ is a distance function of a point to a line segments in 2D, and $\hat d > 0$ is a threshold parameter. In practice, position and orientation sampling is done by rejection sampling and bin sampling (bin size of 6), respectively.

After that, a greedy selector iteratively select the best candidate until $\mathcal{E}^{\mathrm{uncut}}$ is covered. Let $U\leftarrow\mathcal{E}^{\mathrm{uncut}}$ be the current uncover set,
an element $e \in U$ is covered by a candidate $s$ if they intersects (i.e. $s \cap e \neq \varnothing$). Candidates are ranked based on the number of element they can cover. The best candidate is draw as
\begin{equation}
s^\star \in 
\arg\max_{s \in S}\;
\Bigl\lvert \,\bigl\{\, e \in U \;\big|\; s \cap e \neq \varnothing \,\bigr\}\,\Bigr\rvert.
\end{equation}
Then update the selection and uncovered set:
\begin{equation}
\mathcal{G}_{\text{next}} \leftarrow \mathcal{G}_{\text{next}} \cup \{s^\star\},\ 
U \leftarrow U \setminus \bigl\{\, e \in U \;\big|\; s^* \cap e \neq \varnothing \,\bigr\}.
\end{equation}
Finally, $\mathcal{G}_{\text{next}}$ are given back to dissection controller as next dissection goals.

%% file: tex_files/experiment.tex
\begin{figure}[!t]
  \centering
  \begin{subfigure}[b]{1.0\columnwidth}
    \includegraphics[width=\linewidth]{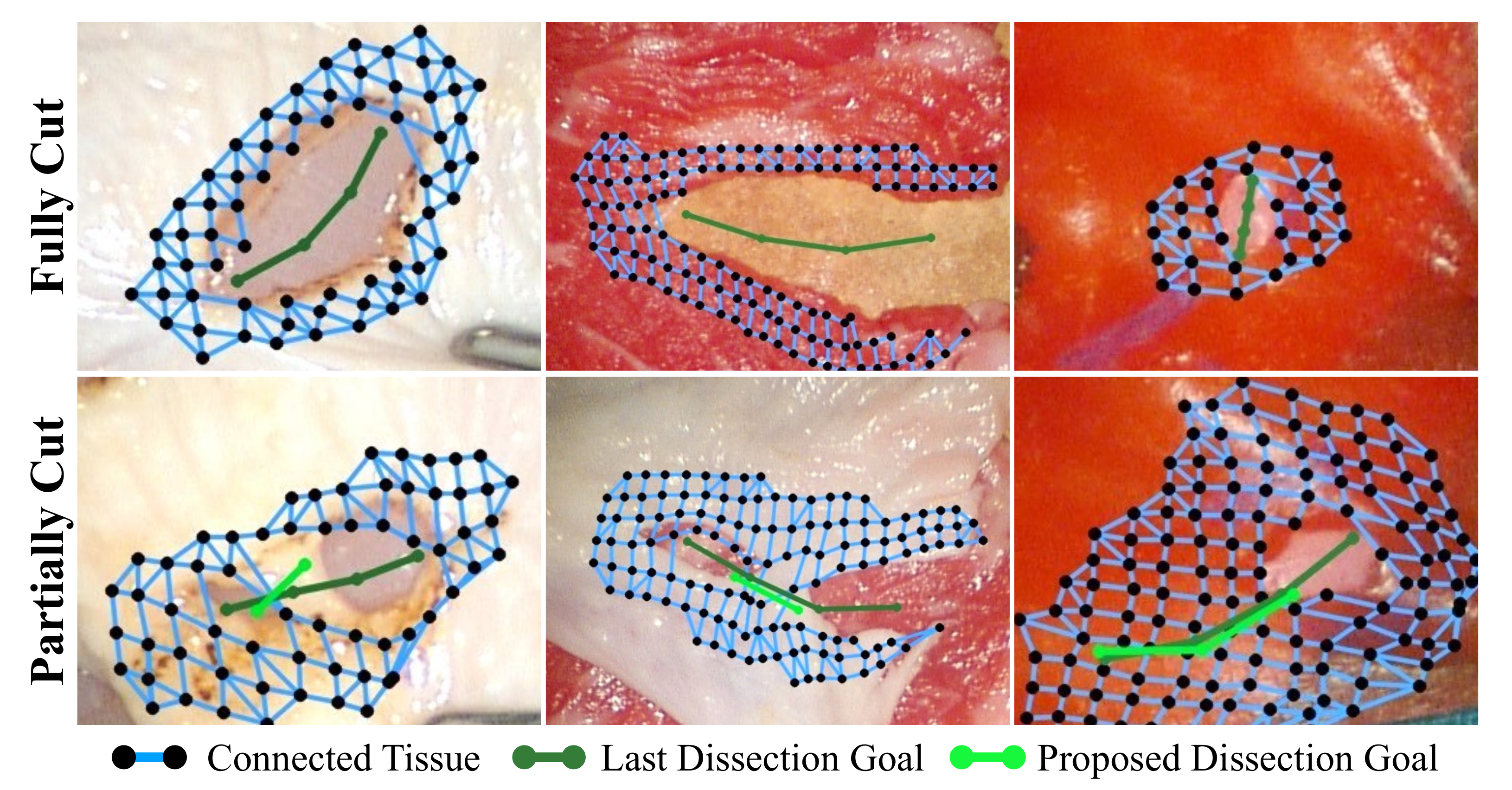}
  \end{subfigure}

  \caption{Qualitative results of the recovery feedback planner across various tissues and conditions. 
  After executing the last dissection goal, each sub-figure shows the resulting tissue connectivity and the planner’s next proposed target.
  }
  \label{fig:dissection_examples}\vspace{-1.5em}
\end{figure}
\section{Experiments}
We conduct extensive real-world experiments and detailed analysis to demonstrate the effectiveness of our method.
All real-world experiments were performed on the da Vinci Research Kit(dVRK) \cite{dvrk}. 
Our algorithm controls two patient-side manipulators (PSM1, PSM2): one equipped with fenestrated bipolar forceps for tissue manipulation and retraction, and the other with pott scissors for dissection. 
To better align with real clinical procedures for the real animal tissue experiment, we replaced the scissors with a cautery spatula tool.

\subsection{Evaluation of Dissection Feedback}
This section evaluates the two components of our dissection feedback framework. 
We first assess the performance of our tissue connectivity estimation across different image keypoint tracking methods.
We then demonstrate the robustness and generalizability of our recovery feedback planning across a diverse set of tissues and conditions.
We curated a dataset of 50 evaluation cases featuring a variety of materials: ex vivo animal tissues (feature-rich beef and feature-poor chicken skin) and silicone tissue phantoms.
Each category includes both fully-cut and partially-cut samples to represent various states of dissection. 
To ensure a consistent and high-quality baseline, all exposure maximizing actions in the dataset were performed under expert human teleoperation.


To quantitatively evaluate the performance of our tissue connectivity estimation, we propose a metric, Detection Accuracy, defined as $\frac{N_{\text{correct}}}{N_{\text{total}}} \times 100\%$,
where $N_{\text{total}}$ is the total number of keypoint pairs evaluated, and $N_{\text{correct}}$ is the number of pairs whose predicted separation state matches the ground-truth (connected or separated) labeled by human.


\subsubsection{Comparison of Different Tracking Methods}
CoTracker3~\cite{cotracker3}, RAFT~\cite{raft}, and LiteTracker~\cite{litetracker} were evaluated for tissue connectivity estimation on our dataset.
LiteTracker achieved the best average detection accuracy (92.8\%), followed by CoTracker3 (90.5\%), while RAFT performed substantially worse (73.4\%).
Given its highest accuracy and lowest computational overhead, we use LiteTracker in the remainder of this work.


\begin{figure}[!t]
  \centering
  \begin{subfigure}[b]{1.0\columnwidth}
    \includegraphics[width=\linewidth]{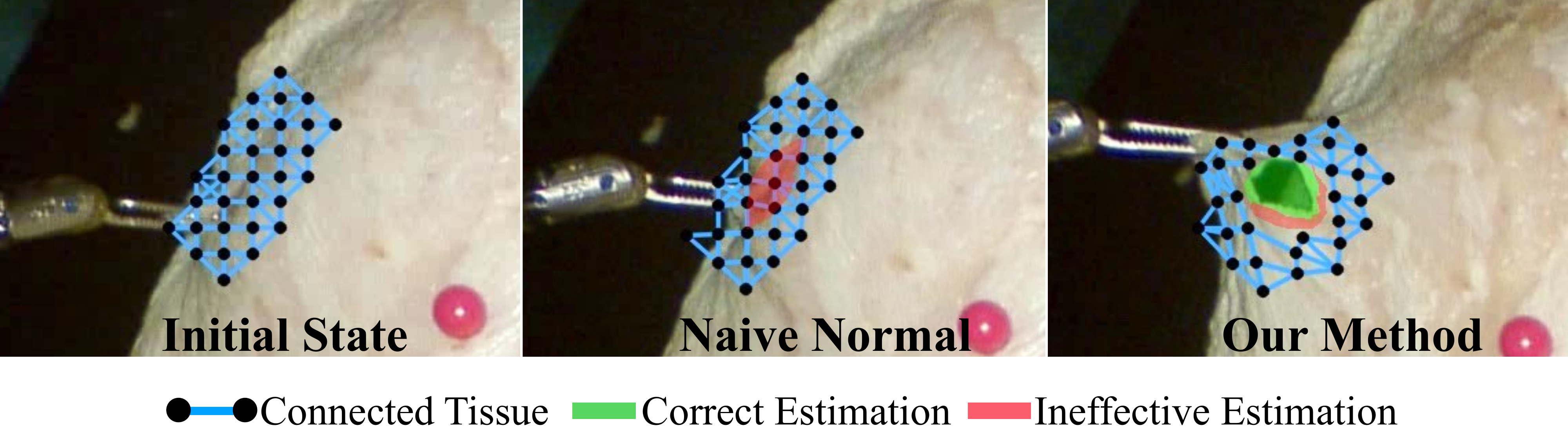}
  \end{subfigure}
  \caption{Comparison of our exposure maximization control and naive normal for improving visibility of the dissected region. For the same amount of retraction, our method achieves significantly more exposure of the dissected area.}
  \label{fig:reveal_bs_ours}\vspace{-1.5em}
\end{figure}

\begin{figure*}[t]
  \centering
\includegraphics[width=1\linewidth]{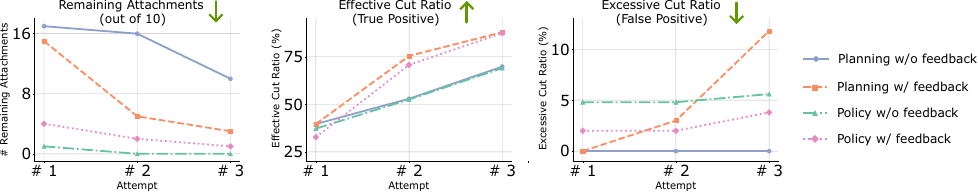}
  \caption{Comparison of dissection performance over three attempts: 
  (a) remaining attachments, (b) effective cut ratio, and (c) excessive cut ratio.}
  \label{fig:cut-results}\vspace{-2em}
\end{figure*}

\subsubsection{Quality of Recovery Feedback Planner}
We demonstrate the robustness and generalizability of our recovery feedback planner by testing it across diverse conditions, including different tissues (ex vivo beef and chicken skin), dissection states (partially and fully cut), and surgical instruments. 
The tools used were a mechanical cutter (Pott's scissors) and an energy-based device (cautery spatula). Fig.~\ref{fig:dissection_examples}  showcases the high quality of the dissection goals generated by our method in these scenarios.

\subsection{Evaluation of Exposure Maximization Control}
In this section, we evaluate the effectiveness of our exposure maximization control method. 
We conduct experiments on real animal tissue for 20 times to demonstrate that our control method not only enhances camera visibility but also improves the performance of the tissue connectivity estimation.
For each trial, we place a real chicken tissue on a $20^\circ$ slope with a predefined dissection goal and an initial dissected region on the chicken skin. 
We then employ our exposure maximization control to manipulate the tissue to maximize the camera visibility of this dissected region. 

To quantify visibility, we manually contour the dissected region in the camera image and count the number of visible pixels on an $480\times640$ image. 
To assess the downstream benefit of improved visibility, we use a connectivity estimation accuracy defined as $\frac{N_{\text{pred}}}{N_{\text{GT}}} \times 100\%$. 
Here, $N_{\text{pred}}$ is the number of edges predicted as dissected (i.e., separated), and $N_{\text{GT}}$ is the ground-truth number of dissected edges labeled by a human expert.

\textbf{Effects of Exposure Maximization Control.}
We analyze the effectiveness of our control method by comparing it against a naive normal planner.
This baseline method plans a straight line trajectory in the direction of the normal vector computed from the grasping point and the midpoint of the dissection goal. 
For a fair comparison, both methods move a total of 0.5 mm per step for 35 steps.
The results are summarized in Table \ref{tab:revealing-results}. 
With our method, the average visible area increased 23 times compared to the naive planner, indicating that our method successfully increases the exposure of the dissected area on real animal tissue. 
Furthermore, our method achieves a superior connectivity estimation accuracy.
This is attributed to our controller's ability to create more visible deformation around the dissected area, providing richer information to the tissue connectivity estimator, as illustrated in Fig.~\ref{fig:reveal_bs_ours}.
\begin{table}[t]
\setlength\tabcolsep{0.3em}
  \centering
  \caption{Comparison of visible area and connectivity estimation accuracy with and without our exposure maximization control.}
    \begin{tabular}{c|cc}
    \toprule
    & Visible Area (pxs) & Connect. Est. Acc. (\%) \\
    \midrule
    Ours-Normal & $69.38\pm60.68$  & $5.56$ \\
    Ours & $\mathbf{1653.2\pm506.57}$ & $\mathbf{86.00}$ \\
    \bottomrule
    \end{tabular}
  \label{tab:revealing-results}
\end{table}

\subsection{Evaluation of Dissection Autonomy with Feedback}

In this section, we present experiments designed to demonstrate how our feedback framework improves the cutting performance of different autonomous dissection agents by enabling them to recover from errors.
we constructed a scene with a volumetric silicon phantom containing a simulated tumor, which is covered by a second, thin layer of textured phantom tissue. Our goal is to dissect the thin-shell.

We evaluate both autonomous dissection agents presented in \autoref{sec:policies} for 60 times (each 10 times), highlighting the flexibility of our feedback approach.
For each trial, an initial dissection goal is manually labeled on the image, and for the planning-based method, a grasping point is also manually specified.


As quantifying dissection performance in real-world scenarios is challenging, we assess the quality of each dissection using two primary metrics: (1) Length Deviation (mm): The difference between the length of the expected dissection path and the executed.
(2) Remaining Attachments: The total number of uncut tissue strands remaining within the desired dissection region.
Finally, we use these measurements to define a successful cut as Length Deviation of less than 2 mm and has zero Remaining Attachments.

\begin{table*}[t]
  \centering
  \caption{Comparison of dissection performance on tissue phantoms. Here, $n$ denotes the number of execution attempts.}
  \label{tab:policy-feedback}
  \begin{tabular}{lcccc}
    \toprule
    \textbf{Dissection Autonomy} & \textbf{Length Deviation (mm)} \downarrowgreen & \textbf{Remaining Attachments (\#)}  \downarrowgreen & \textbf{ Success Rate (\%)} \uparrowgreen\\
    \midrule
    Planning-based w/o Feedback (n=1)     &   $3.60\pm1.44$    &  $16$    &   $20$ &   \\
    Planning-based w/o Feedback (n=3)     &   $3.40\pm1.19$    &  $10$    &   $0$  &   \\
    Learning-based w/o Feedback (n=1)  &   $5.80\pm2.27$    &  $0$     &   $45.5$ &   \\
    Learning-based w/o Feedback (n=3)  &   $4.19\pm2.98$    &  $0$     &   $40$ &   \\
    Planning-based w Feedback             &   $\mathbf{1.47\pm 0.96}$ &  $3$  &     $\textbf{80}$ &   \\
    Learning-based w Feedback          &   $\mathbf{1.84 \pm 1.51}$ &  $0$  &     $\textbf{80}$ &   \\
    \bottomrule
  \end{tabular}
\end{table*}
\begin{figure*}[t]
  \centering
\includegraphics[width=0.95\linewidth]{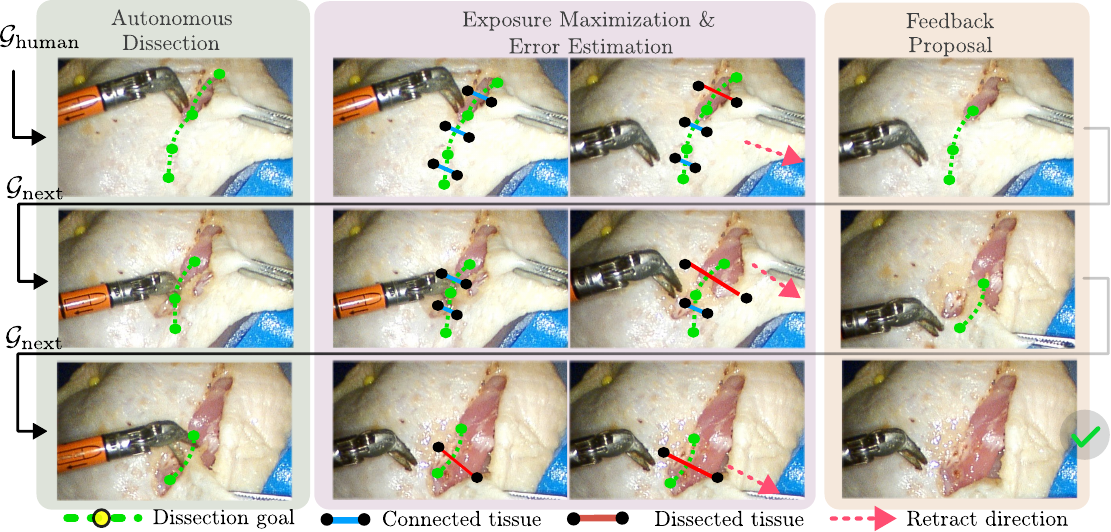}
  \caption{This figure illustrates the execution of our complete feedback-enabled dissection pipeline. \textbf{(First Column)} The process begins with an agent executing a dissection goal. \textbf{(Second Column)} Next, our exposure maximization control retracts the tissue orthogonally to the cut to enhance visibility for error estimation. \textbf{(Third Column)} Then, our feedback planner identifies any remaining tissue attachments and generates a corrective goal. \textbf{(Bottom Right)} This loop repeats for up to three attempts until the dissection is complete. }
  \label{fig:demo-results}
  \vspace{-1em}
\end{figure*}

\textbf{Effects of dissection feedback.} 
We compare three distinct conditions: (1) Ours (w/ Feedback): An autonomous dissection agent uses feedback to refine its dissection goal for up to three attempts.
(2) Baseline (w/o Feedback, n=1): An agent without feedback that executes its dissection goal only once.
(3) Baseline (w/o Feedback, n=3): An agent without feedback that executes the same initial dissection goal for three attempts.
The second baseline was given three attempts to ensure a fair comparison by matching the maximum number of attempts of our feedback-enabled agents.
Their results are summarized in Table~\ref{tab:policy-feedback}, showing that our feedback framework provides a significant performance improvement.
With our dissection feedback framework, both autonomous agents achieve a much higher \textit{Success Rate} than the baselines, demonstrating that our method effectively improves dissection performance across different dissection autonomy paradigms.
Notably, the two agents exhibited distinct failure modes, showcasing the versatility of our feedback mechanism. 
The learning-based agent w/o feedback tended to perform clean cuts that were too short, failing to complete the entire goal. 
Our framework addressed this by identifying the uncut segment and generating a refined goal to complete the dissection.
This recovery is quantitatively reflected in a significant drop in Length Deviation.
In contrast, the planning-based agent was prone to leaving small tissue attachments, especially when the dissection path was long. 
Our framework successfully localized these uncut attachments and provided precise, corrective dissection goals to sever them. 
This is evidenced by a sharp decrease in the number of Remaining Attachments. 
Without feedback, both paradigms performed poorly. 
While blindly re-executing the dissection three times offered a minor improvement over a single attempt, it was insufficient for achieving a clean, successful cut, highlighting the critical importance of our corrective feedback mechanism.
To better understand error correction capabilities of our feedback meachnism, we evaluate the progression of dissection performance over three attempts, with results shown in Fig.~\ref{fig:cut-results}. 
The results show that for the feedback-enabled agent, remaining attachments decrease sharply, the effective cut ratio increases significantly, and excessive cuts are suppressed over the three attempts. 
These trends demonstrate our feedback mechanism's success in correcting imperfect dissections by simultaneously reducing different types of errors.

\subsection{Real Tissue Demonstration}
Finally, we demonstrate our complete framework (planning-based w feedback) on a long-horizon dissection task using ex vivo chicken tissue. 
Our framework successfully separate a 7.6 cm tissue segment sequentially, demonstrating its effectiveness in a realistic, multi-step scenario. Outcomes are shown in Fig. ~\ref{fig:demo-results}.

In the first column, our planning-based dissection agent execute a dissection given a dissection goal. 
Initially the goal was given by a human expert. 
Next our exposure maximization control deforms the tissue to maximize exposure for tissue connectivity estimation (second column). 
In the figure, the retraction direction is orthogonal to the dissection goal, making the deformation difference between the connected tissue and the dissected tissue more visible for better error estimation.
Then, our recovery feedback planner provide a corrective goal as the next goal for the agent to execute.
We limit the dissection attempts to 3 times to finally finish the dissection (bottom right).



\section{Discussion and Conclusion}


In this work, we introduce a feedback-enabled framework for autonomous tissue dissection. 
This framework is designed to reason about and localize topological changes following each dissection action, providing corrective, actionable feedback to the autonomous agent to recover from erroneous dissections. 
To ensure the robustness of this feedback, we propose a novel dissection visibility and an optimal controller that guides the robot to maximize this visibility through active tissue manipulation. 
Through extensive experiments and detailed analysis of each component on both tissue phantoms and ex vivo animal tissues, we demonstrate the effectiveness of our method as a promising step toward robust dissection autonomy in surgery.

The primary limitation of our framework is its focus on thin-shell tissues, as it models and tracks keypoints only on the tissue surface. 
Consequently, the method may struggle to address occluded, partially folded, or volumetric tissues.
However, recognizing that real-world surgical procedures predominantly require volumetric dissection, we seek to expand our feedback mechanism in future work to accommodate volumetric tissues. 
Furthermore, we intend to test the augmented framework on porcine tissue to establish its feasibility in realistic surgical settings.

%% file: root.bib
@article{kim2025srt,
  title={SRT-H: A hierarchical framework for autonomous surgery via language-conditioned imitation learning},
  author={Kim, Ji Woong and Chen, Juo-Tung and Hansen, Pascal and Shi, Lucy Xiaoyang and Goldenberg, Antony and Schmidgall, Samuel and Scheikl, Paul Maria and Deguet, Anton and White, Brandon M and Tsai, De Ru and others},
  journal={Science robotics},
  volume={10},
  number={104},
  pages={eadt5254},
  year={2025},
  publisher={American Association for the Advancement of Science}
}

@article{kim2024surgical,
  title={Surgical robot transformer (srt): Imitation learning for surgical tasks},
  author={Kim, Ji Woong and Zhao, Tony Z and Schmidgall, Samuel and Deguet, Anton and Kobilarov, Marin and Finn, Chelsea and Krieger, Axel},
  journal={arXiv preprint arXiv:2407.12998},
  year={2024}
}

@article{ou2025cressim,
  title={Cressim-mpm: A material point method library for surgical soft body simulation with cutting and suturing},
  author={Ou, Yafei and Tavakoli, Mahdi},
  journal={arXiv preprint arXiv:2502.18437},
  year={2025}
}

@article{williams2015model,
  title={Model predictive path integral control using covariance variable importance sampling},
  author={Williams, Grady and Aldrich, Andrew and Theodorou, Evangelos},
  journal={arXiv preprint arXiv:1509.01149},
  year={2015}
}

@misc{warp2022,
  title        = {Warp: A High-performance Python Framework for GPU Simulation and Graphics},
  author       = {Miles Macklin},
  month        = {March},
  year         = {2022},
  note         = {NVIDIA GPU Technology Conference (GTC)},
  howpublished = {\url{https://github.com/nvidia/warp}}
}

@inproceedings{ge2024enhancing,
  title={Enhancing Surgical Precision in Autonomous Robotic Incisions via Physics-Based Tissue Cutting Simulation},
  author={Ge, Jiawei and Kilmer, Ethan and Mady, Leila J and Opfermann, Justin D and Krieger, Axel},
  booktitle={2024 IEEE/RSJ International Conference on Intelligent Robots and Systems (IROS)},
  pages={2421--2428},
  year={2024},
  organization={IEEE}
}

@ARTICLE{11027660,
  author={Schüßler, Alexander and Kunz, Christian and Younis, Rayan and Alt, Benjamin and Paik, Jamie and Wagner, Martin and Mathis-Ullrich, Franziska},
  journal={IEEE Robotics and Automation Letters}, 
  title={Semi-Autonomous Robotic Assistance for Gallbladder Retraction in Surgery}, 
  year={2025},
  volume={10},
  number={7},
  pages={7468-7475},
  doi={10.1109/LRA.2025.3577430}}

@article{richter2021autonomous,
  title={Autonomous robotic suction to clear the surgical field for hemostasis using image-based blood flow detection},
  author={Richter, Florian and Shen, Shihao and Liu, Fei and Huang, Jingbin and Funk, Emily K and Orosco, Ryan K and Yip, Michael C},
  journal={IEEE Robotics and Automation Letters},
  volume={6},
  number={2},
  pages={1383--1390},
  year={2021},
  publisher={IEEE}
}

@INPROCEEDINGS{10610263,
  author={Liang, Xiao and Liu, Fei and Zhang, Yutong and Li, Yuelei and Lin, Shan and Yip, Michael},
  booktitle={2024 IEEE International Conference on Robotics and Automation (ICRA)}, 
  title={Real-to-Sim Deformable Object Manipulation: Optimizing Physics Models with Residual Mappings for Robotic Surgery}, 
  year={2024},
  volume={},
  number={},
  pages={15471-15477},
  doi={10.1109/ICRA57147.2024.10610263}}

@article{chiu2022real,
  title={Real-time constrained 6d object-pose tracking of an in-hand suture needle for minimally invasive robotic surgery},
  author={Chiu, Zih-Yun and Richter, Florian and Yip, Michael C},
  journal={arXiv preprint arXiv:2210.11973},
  year={2022}
}

@INPROCEEDINGS{medic_ieee,
  author={Liang, Xiao and Wang, Chung-Pang and Shinde, Nikhil Uday and Liu, Fei and Richter, Florian and Yip, Michael},
  booktitle={2025 IEEE International Conference on Robotics and Automation (ICRA)}, 
  title={MEDiC: Autonomous Surgical Robotic Assistance to Maximizing Exposure for Dissection and Cautery}, 
  year={2025},
  volume={},
  number={},
  pages={6889-6895},
  doi={10.1109/ICRA55743.2025.11128739}}

@misc{medic_arxiv,
      title={MEDiC: Autonomous Surgical Robotic Assistance to Maximizing Exposure for Dissection and Cautery}, 
      author={Xiao Liang and Chung-Pang Wang and Nikhil Uday Shinde and Fei Liu and Florian Richter and Michael Yip},
      year={2024},
      eprint={2409.14287},
      archivePrefix={arXiv},
      primaryClass={cs.RO},
      url={https://arxiv.org/abs/2409.14287}, 
}

@misc{lucas_Differentiable_tool_tracking,
      title={Differentiable Rendering-based Pose Estimation for Surgical Robotic Instruments}, 
      author={Zekai Liang and Zih-Yun Chiu and Florian Richter and Michael C. Yip},
      year={2025},
      eprint={2503.05953},
      archivePrefix={arXiv},
      primaryClass={cs.RO},
      url={https://arxiv.org/abs/2503.05953}, 
}

@INPROCEEDINGS{10610378,
  author={D’Ambrosia, Christopher and Richter, Florian and Chiu, Zih-Yun and Shinde, Nikhil and Liu, Fei and Christensen, Henrik I. and Yip, Michael C.},
  booktitle={2024 IEEE International Conference on Robotics and Automation (ICRA)}, 
  title={Robust Surgical Tool Tracking with Pixel-based Probabilities for Projected Geometric Primitives}, 
  year={2024},
  volume={},
  number={},
  pages={15455-15462},
  doi={10.1109/ICRA57147.2024.10610378}}

@INPROCEEDINGS{dvrk,
  author={Kazanzides, Peter and Chen, Zihan and Deguet, Anton and Fischer, Gregory S. and Taylor, Russell H. and DiMaio, Simon P.},
  booktitle={2014 IEEE International Conference on Robotics and Automation (ICRA)}, 
  title={An open-source research kit for the da Vinci® Surgical System}, 
  year={2014},
  volume={},
  number={},
  pages={6434-6439},
  doi={10.1109/ICRA.2014.6907809}}

@inproceedings{franccois2021image,
  title={Image-based incision detection for topological intraoperative 3D model update in augmented reality assisted laparoscopic surgery},
  author={Fran{\c{c}}ois, Tom and Calvet, Lilian and S{\`e}ve-d’Erceville, Callyane and Bourdel, Nicolas and Bartoli, Adrien},
  booktitle={Medical Image Computing and Computer Assisted Intervention--MICCAI 2021: 24th International Conference, Strasbourg, France, September 27--October 1, 2021, Proceedings, Part IV 24},
  pages={647--656},
  year={2021},
  organization={Springer}
}

@article{marahrens2024ultrasound,
  title={An Ultrasound-guided System for Autonomous Marking of Tumor Boundaries during Robot-assisted Surgery},
  author={Marahrens, N and Jones, D and Murasovs, N and Biyani, CS and Valdastri, P},
  journal={IEEE Transactions on Medical Robotics and Bionics},
  year={2024},
  publisher={Institute of Electrical and Electronics Engineers}
}

@article{karimi2024reward,
  title={Reward Learning from Suboptimal Demonstrations with Applications in Surgical Electrocautery},
  author={Karimi, Zohre and Ho, Shing-Hei and Thach, Bao and Kuntz, Alan and Brown, Daniel S},
  journal={arXiv preprint arXiv:2404.07185},
  year={2024}
}

@inproceedings{ge2021supervised,
  title={Supervised autonomous electrosurgery for soft tissue resection},
  author={Ge, Jiawei and Saeidi, Hamed and Kam, Michael and Opfermann, Justin and Krieger, Axel},
  booktitle={2021 IEEE 21st International Conference on Bioinformatics and Bioengineering (BIBE)},
  pages={1--7},
  year={2021},
  organization={IEEE}
}

@article{oh2023framework,
  title={A framework for automated dissection along tissue boundary},
  author={Oh, Ki-Hwan and Borgioli, Leonardo and Zefran, Milos and Chen, Liaohai and Giulianotti, Pier Cristoforo},
  journal={arXiv preprint arXiv:2310.09669},
  year={2023}
}

@article{shinde2024jiggle,
  title={JIGGLE: An Active Sensing Framework for Boundary Parameters Estimation in Deformable Surgical Environments},
  author={Shinde, Nikhil Uday and Liang, Xiao and Liu, Fei and Zhang, Yutong and Richter, Florian and Herbert, Sylvia and Yip, Michael C},
  journal={Conference on Robotics: Science and Systems (RSS)},
  year={2024}
}

@article{liang2023real,
  title={Real-to-Sim Deformable Object Manipulation: Optimizing Physics Models with Residual Mappings for Robotic Surgery},
  author={Liang, Xiao and Liu, Fei and Zhang, Yutong and Li, Yuelei and Lin, Shan and Yip, Michael},
  journal={arXiv preprint arXiv:2309.11656},
  year={2023}
}

@ARTICLE{10354422,
  author={Ge, Jiawei and Kam, Michael and Opfermann, Justin D. and Saeidi, Hamed and Leonard, Simon and Mady, Leila J. and Schnermann, Martin J. and Krieger, Axel},
  journal={IEEE Robotics and Automation Letters}, 
  title={Autonomous System for Tumor Resection (ASTR) - Dual-Arm Robotic Midline Partial Glossectomy}, 
  year={2024},
  volume={9},
  number={2},
  pages={1166-1173},
  doi={10.1109/LRA.2023.3341773}}

@ARTICLE{8884158,
  author={Saeidi, H. and Ge, J. and Kam, M. and Opfermann, J. D. and Leonard, S. and Joshi, A. S. and Krieger, A.},
  journal={IEEE Transactions on Medical Robotics and Bionics}, 
  title={Supervised Autonomous Electrosurgery via Biocompatible Near-Infrared Tissue Tracking Techniques}, 
  year={2019},
  volume={1},
  number={4},
  pages={228-236},
  doi={10.1109/TMRB.2019.2949870}}

@InProceedings{10.1007/978-3-030-32239-7_36,
author="Ge, Jiawei
and Saeidi, Hamed
and Opfermann, Justin D.
and Joshi, Arjun S.
and Krieger, Axel",
editor="Shen, Dinggang
and Liu, Tianming
and Peters, Terry M.
and Staib, Lawrence H.
and Essert, Caroline
and Zhou, Sean
and Yap, Pew-Thian
and Khan, Ali",
title="Landmark-Guided Deformable Image Registration for Supervised Autonomous Robotic Tumor Resection",
booktitle="Medical Image Computing and Computer Assisted Intervention -- MICCAI 2019",
year="2019",
publisher="Springer International Publishing",
address="Cham",
pages="320--328",
isbn="978-3-030-32239-7"
}

@article{video_depth_anything,
    title={Video Depth Anything: Consistent Depth Estimation for Super-Long Videos},
    author={Chen, Sili and Guo, Hengkai and Zhu, Shengnan and Zhang, Feihu and Huang, Zilong and Feng, Jiashi and Kang, Bingyi},
    journal={arXiv:2501.12375},
    year={2025}
  }

@article{depth_anything_v2,
  title={Depth Anything V2},
  author={Yang, Lihe and Kang, Bingyi and Huang, Zilong and Zhao, Zhen and Xu, Xiaogang and Feng, Jiashi and Zhao, Hengshuang},
  journal={arXiv:2406.09414},
  year={2024}
}

@InProceedings{stereo_anywhere,
    author    = {Bartolomei, Luca and Tosi, Fabio and Poggi, Matteo and Mattoccia, Stefano},
    title     = {Stereo Anywhere: Robust Zero-Shot Deep Stereo Matching Even Where Either Stereo or Mono Fail},
    booktitle = {Proceedings of the Computer Vision and Pattern Recognition Conference (CVPR)},
    month     = {June},
    year      = {2025},
    pages     = {1013-1027}
}

@inproceedings{cotracker,
  title     = {CoTracker: It is Better to Track Together},
  author    = {Nikita Karaev and Ignacio Rocco and Benjamin Graham and Natalia Neverova and Andrea Vedaldi and Christian Rupprecht},
  booktitle = {Proc. {ECCV}},
  year      = {2024}
}

@InProceedings{cotracker3,
    author    = {Nikita Karaev and Iurii Makarov and Jianyuan Wang and Natalia Neverova and Andrea Vedaldi and Christian Rupprecht},
    title     = {{CoTracker3}: Simpler and Better Point Tracking by Pseudo-Labelling Real Videos},
    journal   = {arxiv},
    year      = {2024}
  }

@misc{raft,
      title={RAFT: Recurrent All-Pairs Field Transforms for Optical Flow}, 
      author={Zachary Teed and Jia Deng},
      year={2020},
      eprint={2003.12039},
      archivePrefix={arXiv},
      primaryClass={cs.CV},
      url={https://arxiv.org/abs/2003.12039}, 
}

@misc{litetracker,
      title={LiteTracker: Leveraging Temporal Causality for Accurate Low-latency Tissue Tracking}, 
      author={Mert Asim Karaoglu and Wenbo Ji and Ahmed Abbas and Nassir Navab and Benjamin Busam and Alexander Ladikos},
      year={2025},
      eprint={2504.09904},
      archivePrefix={arXiv},
      primaryClass={cs.CV},
      url={https://arxiv.org/abs/2504.09904}, 
}

@INPROCEEDINGS{ACT, 
    AUTHOR    = {Tony Z. Zhao AND Vikash Kumar AND Sergey Levine AND Chelsea Finn}, 
    TITLE     = {{Learning Fine-Grained Bimanual Manipulation with Low-Cost Hardware}}, 
    BOOKTITLE = {Proceedings of Robotics: Science and Systems}, 
    YEAR      = {2023}, 
    ADDRESS   = {Daegu, Republic of Korea}, 
    MONTH     = {July}, 
    DOI       = {10.15607/RSS.2023.XIX.016} 
}
